\documentclass[conference,a4paper]{IEEEtran}

\IEEEoverridecommandlockouts
\usepackage{cite}
\usepackage{amsmath,amssymb,amsfonts}
\usepackage{algorithmic}
\usepackage{graphicx}
\usepackage[table,xcdraw]{xcolor}
\usepackage{textcomp}
\usepackage{xcolor}
\usepackage{lipsum}
\usepackage{hyperref}
\usepackage{float} 

\usepackage{booktabs} 
\usepackage[table]{xcolor} 
\def\BibTeX{{\rm B\kern-.05em{\sc i\kern-.025em b}\kern-.08em
    T\kern-.1667em\lower.7ex\hbox{E}\kern-.125emX}}

\usepackage{multirow}

\begin{document}

\title{Can Large Language Models Predict the Outcome of Judicial Decisions?}

\author{
    \IEEEauthorblockN{
        Mohamed Bayan Kmainasi\IEEEauthorrefmark{1}\IEEEauthorrefmark{2}
    }
    \IEEEauthorblockA{
        \textit{Computer Science and Engineering Dept.} \\
        \textit{Qatar University}\\
        Doha, Qatar \\
        mk2314890@qu.edu.qa
    }
    \and
    \IEEEauthorblockN{
        Ali Ezzat Shahroor\IEEEauthorrefmark{1}\IEEEauthorrefmark{2}
    }
    \IEEEauthorblockA{
        \textit{School of Computing and Data Science} \\
        \textit{OUC in partnership with LJMU}\\
        Doha, Qatar \\
        100230@oryx.edu.qa
    }
    \and
    \IEEEauthorblockN{
        Amani Al-Ghraibah
    }
    \IEEEauthorblockA{
        \textit{Biomedical Engineering Dept.} \\
\textit{Jordan University of} \\
\textit{Science and Technology}\\
        Irbid, Jordan \\
        \empty
    }
    
    \thanks{\IEEEauthorrefmark{1}These authors contributed equally to this work.}
    \thanks{\IEEEauthorrefmark{2}The contributions were made while the authors were individual contributors at the Qatar Computing Research Institute (QCRI).}
}

\maketitle

\begin{abstract}
Large Language Models (LLMs) have shown exceptional capabilities in Natural Language Processing (NLP) across diverse domains. However, their application in specialized tasks such as Legal Judgment Prediction (LJP) for low-resource languages like Arabic remains underexplored. In this work, we address this gap by developing an Arabic LJP dataset, collected and preprocessed from Saudi commercial court judgments. We benchmark state-of-the-art open-source LLMs, including LLaMA-3.2-3B and LLaMA-3.1-8B, under varying configurations such as zero-shot, one-shot, and fine-tuning using LoRA. Additionally, we employed a comprehensive evaluation framework that integrates both quantitative metrics (such as BLEU, ROUGE, and BERT) and qualitative assessments (including Coherence, Legal Language, Clarity, etc.) using an LLM. Our results demonstrate that fine-tuned smaller models achieve comparable performance to larger models in task-specific contexts while offering significant resource efficiency. Furthermore, we investigate the impact of fine-tuning the model on a diverse set of instructions, offering valuable insights into the development of a more human-centric and adaptable LLM. We have made the dataset, code, and models publicly available to provide a solid foundation for future research in Arabic legal NLP\footnote{https://github.com/MohamedBayan/Arabic-Legal-Judgment-Prediction}.

\end{abstract}

\begin{IEEEkeywords}
Natural language Processing, Legal Judgment Prediction, Large Language Models
\end{IEEEkeywords}

\section{Introduction} 

LLMs have transformed NLP, achieving state-of-the-art (SOTA) performance across diverse tasks in multilingual and multitask settings \cite{zhao2023survey}. Advanced LLMs like GPT-4 and Gemini demonstrate remarkable reasoning, comprehension, and problem-solving capabilities, enabling applications in specialized domains such as law \cite{padiu2024extent}, medicine \cite{liu2024performance, aydin2024large}, and education \cite{imran2024google}.

However, their closed-source nature raises concerns about data privacy, customization, and accessibility, especially for organizations seeking domain-specific solutions \cite{minaee2024large, von2024systematic}. Open-source models like LLaMA-3 \cite{grattafiori2024llama3herdmodels} and Phi-4 \cite{abdin2024phi4technicalreport} provide flexible alternatives, offering fine-tuning and in-context learning capabilities that often rival proprietary systems and exceed baseline results \cite{abdelali-etal-2024-larabench, ding2023parameter}.

A key candidate application of LLMs is LJP, which predicts judicial outcomes based on factual case details \cite{shui2023comprehensiveevaluationlargelanguage}. Currently, legal experts rely on extensive training to identify relevant laws, determine charges, and issue judgments \cite{gordon2014role}. Legal decision-making is inherently complex, requiring analysis of vast case-specific information while ensuring consistency and fairness \cite{baez2020impact}. This process remains labor-intensive, time-consuming, and prone to bias \cite{baez2020impact}.

Advances in NLP are automating aspects of LJP, enabling efficient and consistent data-driven predictions \cite{shui2023comprehensiveevaluationlargelanguage}. Early LJP approaches relied on manually extracted features, which were costly and inefficient \cite{santosh-etal-2022-deconfounding}. LLM integration reduces the need for manual feature engineering, improves prediction accuracy, and simplifies complex legal reasoning.

Despite these advancements, LJP remains underexplored in Arabic due to unique linguistic challenges like morphological richness, dialectal variation, and complex syntax \cite{darwish2021panoramic}. LLMs, successful in high-resource languages like English, often underperform in Arabic, especially with Arabic-prompted instructions \cite{kmainasi2024native}. The lack of publicly available Arabic datasets tailored for LJP further exacerbates this gap, hindering model training and evaluation.

To address these challenges, this work develops a tailored solution for Arabic LJP. Specifically, we:

\begin{enumerate} 

\item Collect and preprocess Arabic legal case data from the Saudi commercial court\footnote{https://laws.moj.gov.sa/ar/}, creating a domain-specific dataset. 

\item Prepare a fine-tuning dataset by crafting 75 diverse Arabic instructions of varying lengths, complexity, and styles to enhance model generalizability and simulate human-centric prompting scenarios.

\item Benchmark open-source LLMs, including LLaMA-3.2-3B-instruct and LLaMA-3.1-8B-instruct, under zero-shot, one-shot, and fine-tuning settings using LoRA. 

\item Evaluate performance using traditional metrics like BLEU and ROUGE, complemented by qualitative analysis using LLaMA-3.1-8B-instruct to assess responses against ground truth judgments. \end{enumerate}

This study addresses the lack of Arabic legal NLP datasets by providing a curated dataset, benchmarking results, implementation code, and trained models. Additionally, it introduces an evaluation pipeline combining quantitative and qualitative assessments, offering a robust framework for advancing LJP research in Arabic and other low-resource languages.

\section{Related Work}
In recent years, Legal Judgment Prediction has attracted substantial attention, driven by advancements in NLP techniques \cite{cui2022surveylegaljudgmentprediction}. Early approaches to LJP primarily utilized rule-based systems \cite{kort1957predicting}, relying on predefined legal rules and templates to predict outcomes. These systems were later replaced by traditional ML models with manually crafted features, introducing statistical methods that improved performance \cite{hsieh2021legal}. The emergence of neural-based approaches \cite{chalkidis-etal-2019-neural}, particularly recurrent neural networks (RNNs) \cite{schmidt2019recurrentneuralnetworksrnns} and transformer-based architectures \cite{vaswani2023attentionneed}, revolutionized the field by enabling models to better capture the semantic complexity of legal texts.

Transformer-based models have proven transformative for legal text analysis by modeling intricate semantic and syntactic relationships. The BERT model \cite{kenton2019bert}, along with its specialized versions such as Legal-BERT \cite{chalkidis2020legal}, Legal-RoBERTa \cite{geng2021legal}, CaseLawBERT, and InCaseLawBERT \cite{paul2023pre}, has demonstrated exceptional performance in extracting semantic richness from legal documents. For instance, \cite{imran2023classifying} showcased BERT's effectiveness in classifying European Court of Human Rights cases, while \cite{maqsood2024transformer} fine-tuned BERT models on the VerdictVaultPK dataset of Pakistani legal case judgments, achieving over 20 F1 points higher than traditional classifiers like TF-IDF and logistic regression. Similarly, \cite{nigam2023fact} applied BERT to predict decisions in Indian legal judgments \cite{malik-etal-2021-ildc} and suggested the use of LLMs to address BERT’s context-length limitations.

Moreover, researchers have explored domain-specific adaptations of transformers for LJP in other languages. \cite{ghosh2024evaluating} demonstrated the importance of specialized transformer models for handling long-form Romanian legal judgments, emphasizing that general-purpose models often lack the contextual depth and domain knowledge required for legal tasks. In criminal law, \cite{latisha2024criminal} applied optimized BERT variants, outperforming sparse text representation models in predicting criminal case outcomes. Despite these advancements, challenges in multilingual and low-resource legal NLP persist. \cite{huang2023not} noted that LLMs excel in high-resource languages like English but struggle with low-resource, non-Latin languages, a finding corroborated by \cite{nguyen2023democratizing}.

Interestingly, English-written instructions often perform strongly even for non-English tasks. For example, \cite{kmainasi2024native} found that non-native English instructions outperformed native Arabic ones in certain Arabic NLP tasks, demonstrating the adaptability of LLMs when leveraging high-quality, generalized instruction data in low-resource settings. Furthermore, \cite{kmainasi2024llamalens} demonstrated that Llama3's performance improved significantly when fine-tuned on high-quality Arabic datasets specifically curated for social media tasks.

Additionally, \cite{ammar2024prediction} utilized LLaMA, GPT, and Jais for Arabic LJP, showing that translating legal cases into English significantly improved model performance. Their findings further emphasize the need for Arabic-centric models tailored to the legal domain.

Recently, the potential of LLMs for LJP has been further explored. \cite{vats2023llms} and \cite{nigam-etal-2024-legal} benchmarked GPT-3.5 and LLaMA-2 on Indian court datasets, while \cite{bommarito2022gpttakesbarexam} evaluated GPT-3’s performance on the Uniform Bar Examination, showcasing its capability in legal reasoning and generating contextually relevant answers. Transformer models like LegalBERT and T5 have also excelled in structured text generation, classification, and summarization tasks \cite{chalkidis2020legalbertmuppetsstraightlaw, raffel2023exploringlimitstransferlearning}, further solidifying their role in automating and enhancing legal systems.

Despite significant advancements in legal NLP, Arabic LJP remains underexplored. Challenges such as the morphological complexity of Arabic, syntax variability, and the scarcity of high-quality annotated datasets have hindered progress. Existing Arabic corpora are often fragmented, domain-specific, or poorly annotated, limiting the development of robust models.

\begin{figure*}[!t]
    \centering
    \includegraphics[width=\textwidth]{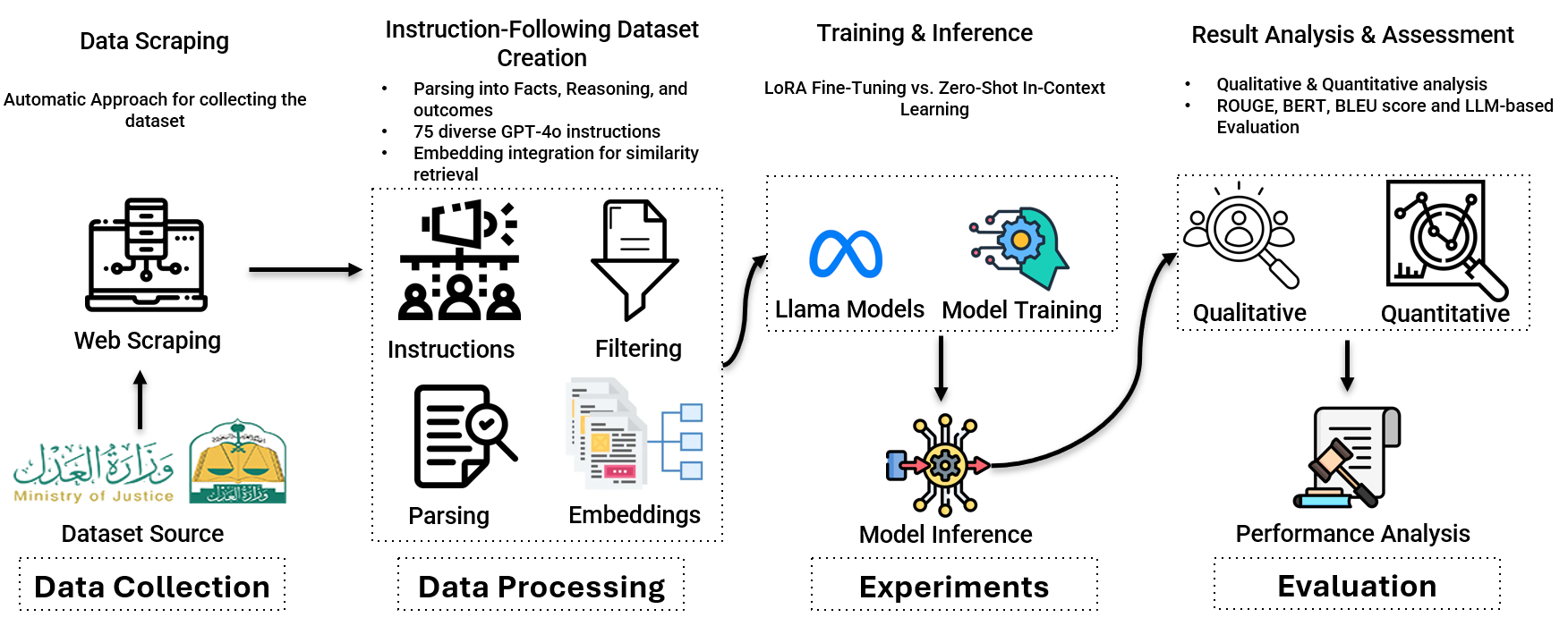}
    \caption{End-to-End Workflow: From Dataset Collection to Evaluation of LLama Models}
    \label{fig:methodology}
\end{figure*}


\section{Methodology}
This section provides a detailed overview of the dataset construction process, the models used, the prompting strategies implemented, and the fine-tuning approach adopted. Fig. \ref{fig:methodology} illustrates the complete end-to-end workflow of our study.

\subsection{Dataset Collection and Preparation}
\begin{figure}
    \centering
    \includegraphics[width=1.\linewidth]{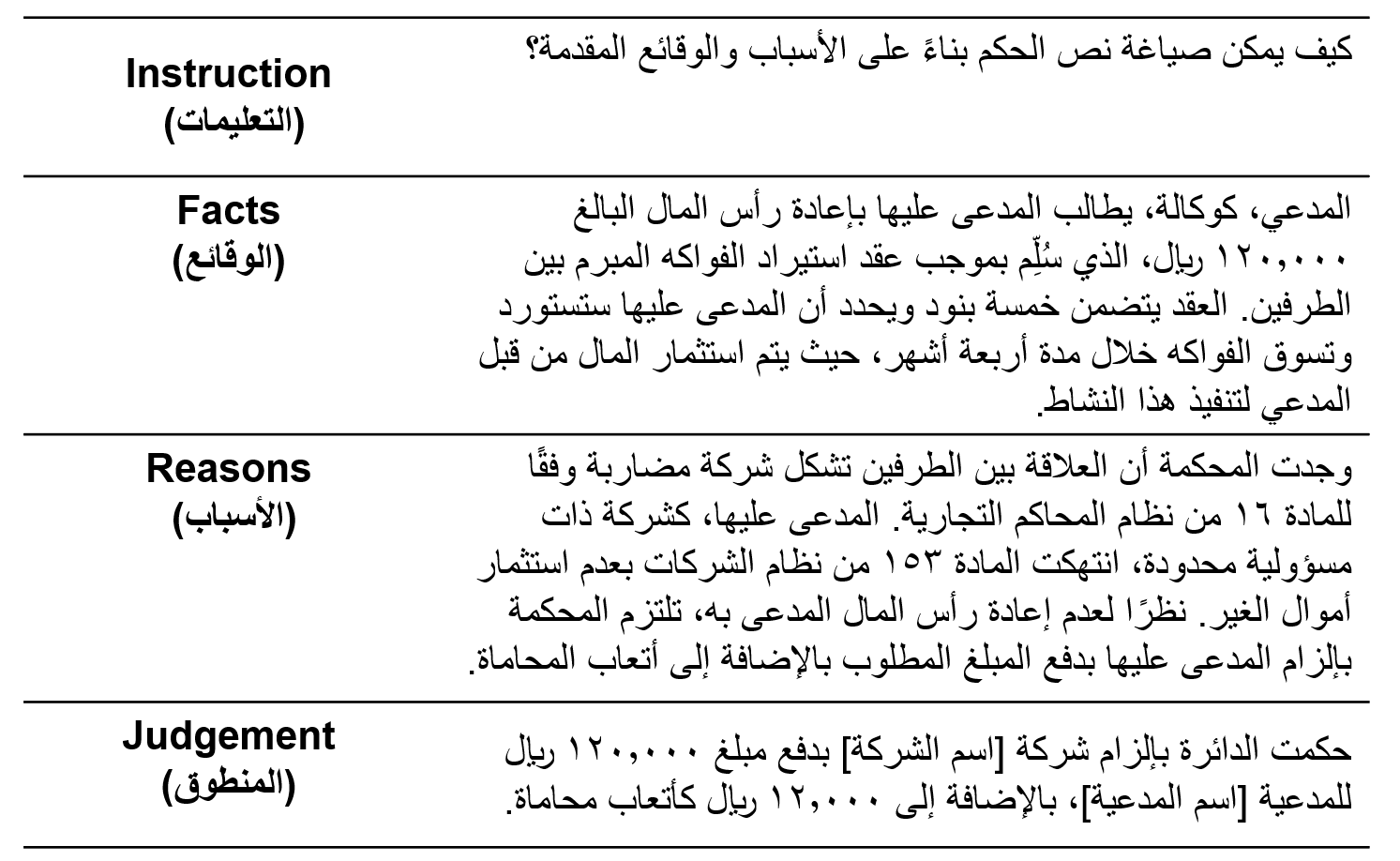}
    \caption{Dataset Sample}
    \label{fig:data_point}
\end{figure}

In this research, we constructed a domain-specific dataset for Arabic LJP by collecting court judgments from the publicly accessible Saudi Ministry of Justice Judgment Publication Platform\footnote{\url{https://laws.moj.gov.sa/ar/}}. This platform hosts a substantial repository of legal cases, particularly within the domain of commercial law. Fig. \ref{fig:data_point} presents a sample data point.

To compile the dataset, we employed web scraping techniques to extract judgments, followed by rigorous parsing and preprocessing steps to organize the data into a structured format. The resulting dataset facilitates the generation of diverse instruction-based examples suitable for fine-tuning language models.

After shuffling the dataset to ensure randomization, we sampled a smaller version of the dataset and split it into training (3752 samples) and testing (538 samples) sets for training and evaluation.
We created 75 diverse Arabic instructions using GPT-4o varying in length and complexity. These instructions were uniformly distributed across the dataset's data points to ensure broad coverage.

The embeddings were generated using a BERT-based sentence embedding model from Sentence Transformers \cite{reimers2019sentencebertsentenceembeddingsusing}, specifically the paraphrase-multilingual-MiniLM-L12-v2 model\footnote{\url{https://huggingface.co/sentence-transformers/paraphrase-multilingual-MiniLM-L12-v2}}. This model supports multiple languages, including Arabic, and enables efficient semantic retrieval of relevant legal judgments.

This following table illustrates how the model processes legal instructions, facts, and reasons to predict judicial outcomes. It highlights the importance of clear prompts and structured data in enhancing LLM performance for Arabic LJP.

\subsection{Fine-Tuning}
We use \textbf{LoRA} (Low-Rank Adaptation) \cite{hu2021loralowrankadaptationlarge}, a parameter-efficient fine-tuning method enabling large pre-trained models to adapt with minimal computational cost. LoRA lowers memory usage while maintaining performance, allowing fine-tuning on resource-limited hardware for large-scale tasks.



Both models were trained for two epochs on a single A100 GPU, using a batch size of 32 and a maximum sequence length of 2048. LoRA was applied with a rank of \( r = 64 \), a scaling factor of \( \alpha = 64 \), and a dropout rate of 0.1. The target modules included \texttt{q\_proj}, \texttt{k\_proj}, \texttt{v\_proj}, \texttt{o\_proj}, \texttt{gate\_proj}, \texttt{up\_proj}, and \texttt{down\_proj}, all with a dropout of 0.1. Training was performed for one epoch using the AdamW optimizer \cite{kingma2017adammethodstochasticoptimization} with a learning rate of \( 2 \times 10^{-4} \).

\subsection{Evaluation Metrics}
We evaluated the model using both \textbf{quantitative} and \textbf{qualitative} metrics.  

For \textbf{quantitative evaluation}, we employed ROUGE-F \cite{lin2004rouge}, BLEU \cite{papineni2002bleu}, and BERTScore \cite{zhang2020bertscoreevaluatingtextgeneration} which are standard metrics for text generation tasks:  
\begin{itemize}  
    \item \textbf{ROUGE-F}: Measures the F1-score (harmonic mean of precision and recall) for overlapping n-grams, capturing content similarity and informativeness between the generated text and the reference text.
    \item \textbf{BLEU}: Evaluates n-gram precision with a brevity penalty, assessing fluency and alignment with the reference.
    \item \textbf{BERT-score-F}: Computes the F1-score using contextual embeddings to measure semantic similarity, making it well-suited for LJP, where precise legal understanding is crucial.
\end{itemize}  

For \textbf{qualitative evaluation}, we used the \textbf{LLaMA3.1-8B-Instruct} model to rate the generated responses against the ground truth on a scale of 1 to 10 across eight dimensions:  
\begin{enumerate}  
    \item \textbf{Accuracy:} Alignment with factual and legal details.  
    \item \textbf{Relevance:} Appropriateness to the legal question.  
    \item \textbf{Coherence:} Logical organization and consistency.  
    \item \textbf{Brevity:} Conciseness with sufficient detail.  
    \item \textbf{Legal Language:} Use of formal legal terminology.  
    \item \textbf{Faithfulness:} Preservation of ground truth facts and principles.  
    \item \textbf{Clarity:} Ease of understanding.  
    \item \textbf{Consistency:} Absence of contradictions.  
\end{enumerate}  
Our use of LLMs as evaluators is inspired by previous studies \cite{hada-etal-2024-large, chiang2023can}.

\subsection{Prompting Strategies}  

\subsubsection{Zero-shot Prompting}  
In zero-shot prompting, the LLM generates a legal judgment based on an instruction and the provided case details. Formally:  

\[
f(\text{Instruction}, \text{Facts + Reasons}) \rightarrow \text{Judgment}
\]  

\noindent where:  
\begin{itemize}  
    \item \textbf{\(f\)}: The LLM.  
    \item \textbf{\(\text{Instruction}\)}: A prompt specifying the task, e.g., \textit{"Based on the facts, analyze the reasons and extract the final judgment text."}  
    \item \textbf{\(\text{Facts + Reasons}\)}: The factual circumstances of the case, along with the legal reasoning and principles applied.  
    \item \textbf{\(\text{Judgment}\)}: The model’s output decision.  
\end{itemize}  

\subsubsection{Few-shot Prompting}  
Few-shot prompting provides the LLM with examples from the training data to guide its judgment. In our case, we used only a single example. The formalization remains the same as in zero-shot prompting, with the addition of an example:

\[
f(\text{Instruction}, \text{Example}, \text{Facts + Reasons}) \rightarrow \text{Judgment}
\]  

\noindent where:  
\begin{itemize}  
    \item \textbf{\(\text{Example}\)}: A case from the training set, represented as:  
    \[
    \text{Example} = (\text{Facts + Reasons}_{\text{ex}}, \text{Judgment}_{\text{ex}})
    \]  
\end{itemize}  

The example guides the model by demonstrating the relationship between input (\(\text{Facts + Reasons}_{\text{ex}}\)) and output (\(\text{Judgment}_{\text{ex}}\)), improving alignment with the task. It is retrieved based on semantic similarity to the current case, ensuring relevance in the legal reasoning applied.

\subsection{Model Inference}
The models were deployed for inference using \texttt{vLLM} \cite{kwon2023efficientmemorymanagementlarge} on four NVIDIA A16 GPUs. The maximum sequence length was set to 2048 tokens. The temperature parameter was fixed at zero to ensure reproducibility and deterministic outputs.

\section{Experiments and Results}

Table~\ref{tab:llama_comparison} presents a comparison of the performance of smaller and larger versions of LLaMA, evaluated under 1-shot learning and fine-tuning settings, across both subjective and objective evaluation metrics.

\subsection{LLM-Based Evaluation Analysis}

Fine-tuning led to a substantial improvement in objective performance across all evaluated metrics. For the smaller \textbf{Llama-3.2-3B} model, fine-tuning resulted in an average score increase from \textbf{3.01 to 7.13}, yielding a \textbf{4.14-point improvement} across all metrics. In comparison, under in-context learning settings, the \textbf{1-shot performance} of Llama-3.2-3B outperformed the baseline by an \textbf{average of 1.65 points}.

For the larger \textbf{Llama-3.1-8B} model, fine-tuning yielded a \textbf{1.55-point improvement}, surpassing both its untrained counterpart and the fine-tuned Llama-3.2-3B. Across all qualitative evaluation criteria—including \textbf{Coherence, Brevity, Legal Language, Faithfulness, Clarity, and Consistency}—both in-context learning and fine-tuning consistently outperformed the untrained models, demonstrating the effectiveness of these adaptation strategies.

\subsection{Standard Evaluation Analysis}

The base models exhibited poor performance when evaluated on standard metrics. For instance, \textbf{Llama-3.2-3B} achieved \textbf{ROUGE-1: 0.08, ROUGE-2: 0.02, BLEU: 0.01,} and \textbf{BERTScore: 0.54}. The larger \textbf{Llama-3.1-8B} performed slightly better, with \textbf{ROUGE-1: 0.12, ROUGE-2: 0.04, BLEU: 0.02,} and \textbf{BERTScore: 0.58}, highlighting the inherent complexity of the task.

\textbf{1-shot learning} significantly improved performance, yielding an absolute increase of \textbf{6\% in BERTScore} (0.64), \textbf{9\% in BLEU}, and \textbf{15\% in ROUGE-2}. Our \textbf{fine-tuned models} demonstrated comparable performance, with an average difference of less than \textbf{3\%}, favoring the larger \textbf{Llama-3.1-8B-FT}. Our best-performing model achieved a \textbf{BERTScore of 0.76}, indicating strong similarity and highlighting the model’s robustness. Similarly, it attained a \textbf{BLEU score of 0.26, ROUGE-2 of 0.41,} and \textbf{ROUGE-1 of 0.53}.

\begin{table*}[htbp!]
    \centering
    \begin{tabular}{lccccc}
        \toprule
        \textbf{Metric} & \textbf{LLaMA-3.2-3B} & \textbf{LLaMA-3.1-8B} & \textbf{LLaMA-3.2-3B-1S} & \textbf{LLaMA-3.2-3B-FT} & \textbf{LLaMA-3.1-8B-FT} \\
        \midrule
        Coherence               & 2.69 & 5.49 & 4.52 & \underline{6.60} & \textbf{6.94} \\
        Brevity                 & 1.99 & 4.30 & 3.76 & \underline{5.87} & \textbf{6.27} \\
        Legal Language          & 3.66 & 6.69 & 5.18 & \underline{7.48} & \textbf{7.73} \\
        Faithfulness            & 3.00 & 5.99 & 4.00 & \underline{6.08} & \textbf{6.42} \\
        Clarity                 & 2.90 & 5.79 & 4.99 & \underline{7.90} & \textbf{8.17} \\
        Consistency             & 3.04 & 5.93 & 5.14 & \underline{8.47} & \textbf{8.65} \\
        Avg. Qualitative Score  & 3.01 & 5.89 & 4.66 & \underline{7.13} & \textbf{7.44} \\
        \midrule
        ROUGE-1                 & 0.08 & 0.12 & 0.29 & \underline{0.50} & \textbf{0.53} \\
        ROUGE-2                 & 0.02 & 0.04 & 0.19 & \underline{0.39} & \textbf{0.41} \\
        BLEU                    & 0.01 & 0.02 & 0.11 & \underline{0.24} & \textbf{0.26} \\
        BERT                    & 0.54 & 0.58 & 0.64 & \underline{0.74} & \textbf{0.76} \\
        \bottomrule
    \end{tabular}
    \vspace{0.5em}
    \caption{A comparative analysis of performance across different LLaMA models. The model names have been abbreviated for simplicity: \textbf{LLaMA-3.2-3B-Instruct} is represented as LLaMA-3.2-3B, \textbf{LLaMA-3.1-8B-Instruct} as LLaMA-3.1-8B, \textbf{LLaMA-3.2-3B-Instruct-1-Shot} as LLaMA-3.2-3B-1S, \textbf{LLaMA-3.2-3B-Instruct-Finetuned} as LLaMA-3.2-3B-FT, and \textbf{LLaMA-3.1-8B-Finetuned} as LLaMA-3.1-8B-FT.}
    \label{tab:llama_comparison}
\end{table*}

\section{Discussion}
This section examines the performance of LLM configurations in LJP, focusing on the effects of instruction generalization, fine-tuning, and in-context learning. We analyze model behaviors, highlight strengths and limitations, and discuss trade-offs in adapting LLMs for this specialized task.

\subsection{Instruction Generalization Analysis}
To evaluate the impact of fine-tuning on model performance and its ability to generalize across a diverse range of instructions, we compared two sets of models: \textbf{Llama-3.2-3b-instruct} and \textbf{Llama-3.2-3b-finetuned}, as well as \textbf{Llama-3.1-8b-instruct} and \textbf{Llama-3.1-8b-finetuned}. The models were trained on a set of 75 diverse instructions to encourage generalization and promote human-centric behavior.

We measured model performance using the BERT F1 score and applied the Wilcoxon signed-rank test to assess statistical significance, with an alpha level set to 0.05. The significance test was applied as follows: the 75 instructions were distributed across the test set, and for each instruction, we computed the BERT F1 score. The individual scores were then aggregated by taking the average BERT score for each model, before and after fine-tuning. The Wilcoxon signed-rank test was then performed on the 75 averaged BERT scores to determine whether the improvements in performance post-fine-tuning were statistically significant.

The results demonstrated significant improvements in performance after fine-tuning. Specifically, the \textbf{Llama-3.2-3b-instruct} model showed a 20\% absolute increase, from 0.54 to 0.64, with a p-value of 0, indicating a statistically significant enhancement. Similarly, the \textbf{Llama-3.1-8b-instruct} model exhibited an 18\% absolute improvement, from 0.58 to 0.74, with a p-value of 0, confirming statistical significance.

These findings underscore the effectiveness of training on a diverse set of instructions to improve instruction generalization and achieve more human-centric performance. The observed improvements in both models after fine-tuning demonstrate the models' enhanced ability to generalize across instructions, while ensuring they remain aligned with human-centric goals.

\subsection{Finetuning vs. In-Context Learning}
As shown in Table~\ref{tab:llama_comparison}, finetning using LoRA significantly outperformed in-context learning across all metrics,
Furthermore, in-context learning requires the model to manage extended context sizes, which becomes increasingly costly in legal domain due to the inherently large and complex nature of legal cases. Moreover, the necessity to embed and store extensive training examples further escalates the computational and storage requirements over time, posing a scalability challenge for practical deployment in real-world legal applications.

Given these considerations, finetuning emerges as the more pragmatic approach. It not only delivers significantly enhanced performance but also maintains a moderate context size, thereby reducing computational overhead and facilitating easier deployment. These findings suggest that a fine-tuning strategy, especially using efficient methods like LoRA, is better suited for advancing legal NLP applications.

\section{Conclusion}

This study introduced the first Arabic instruction-following dataset for LJP and benchmarked LLMs across different learning paradigms. Our results demonstrate that LoRA-based fine-tuning significantly enhances legal NLP performance, outperforming untrained and in-context approaches while reducing computational overhead. We showed the effectiveness of our approach for instruction generalization, to have a human-centric model.

Future work will explore the effectiveness of LLM-based scoring, larger models, and alternative fine-tuning methods. Additionally, developing a multi-domain, multilingual LLM for the legal domain could enhance model versatility and scalability across diverse legal contexts.


\bibliographystyle{IEEEtran}
\bibliography{bibliography}

\end{document}